%% file: main.tex
\pdfoutput=1
\documentclass[11pt]{article}
\usepackage[preprint]{acl}
\usepackage{times}
\usepackage{latexsym}
\usepackage[T1]{fontenc}
\usepackage[utf8]{inputenc}
\usepackage{microtype}
\usepackage{inconsolata}
\usepackage{graphicx}
\usepackage{amsmath}
\usepackage{booktabs}


\makeatletter
\ifacl@anonymize
  \newcommand{\repourl}{\url{https://anonymous.4open.science/r/nli-hlv-selection-A438}}
\else
  \newcommand{\repourl}{\url{https://github.com/oudeis01/nli-hlv-selection}}
\fi
\makeatother

\title{Selection Shapes the Boundary: \\ A Replication of
Monotonicity and Label Agreement in Unselected NLI Populations}

\author{Haram Choi \\
  University of Bremen \\
  \texttt{hachoi@uni-bremen.de}}

\begin{document}
\maketitle

\begin{abstract}
Prior work on human label variation (HLV) in natural language inference
(NLI) has often relied on re-annotation resources that select items by
disagreement level. An earlier study \citep{choi-2026-hlv} found that
hypotheses containing non-upward monotonicity operators showed
somewhat higher label entropy in ChaosNLI (Cliff's $\delta = -0.284$), a
difference that study reports as sensitive to tagger error. In its
SNLI and MultiNLI portions, ChaosNLI re-annotates only items whose
majority label carries exactly three of five votes. We
replicate this boundary in the unselected populations that ChaosNLI was
drawn from. We use the SNLI and MultiNLI development sets with the same
operator tagger and a four-level ordinal agreement outcome, under a plan
frozen in our working repository before analysis. The planned prediction
fails. All seven contrasts return a positive Cliff's $\delta$ (non-upward items agree slightly more,
not less), the only significant confirmatory contrast has the opposite
sign to the planned direction, and every effect is far below our smallest
effect size of interest (0.10). Robustness checks support the measurement: simulated tagger
misclassification shrinks the effects rather than manufacturing them, and
a manual re-tagging audit reaches four-class agreement of 0.875 on a
fresh 200-item sample. We conclude
that the earlier negative boundary is plausibly a structure conditional
on low-agreement selection rather than a population-level property, and
that HLV structure claims built on selected re-annotation resources
should state their selection conditional explicitly.
\end{abstract}

\section{Introduction}
\label{sec:intro}

Human label variation in NLI is increasingly treated as signal rather
than annotation noise: multiple labels per item reflect genuine
interpretive variation that models and evaluations should represent
\citep{plank-2022-problem,pavlick-kwiatkowski-2019-inherent}. Under this
perspectivist frame, disagreement is a measurement target, and a natural
research question is whether linguistic properties of an item predict how
much annotators will disagree on it.

One candidate predictor is monotonicity. \citet{choi-2026-hlv} tagged
hypothesis sentences for monotonicity operators and found that in
ChaosNLI, items whose hypotheses contained non-upward operators
(downward-entailing or non-monotone triggers) had higher label
entropy than purely upward items, with a Cliff's
$\delta$ of $-0.284$ (95\% CI $[-0.332, -0.235]$) on a 100-label
entropy outcome; that study reports the difference as sensitive to
tagger error.

That estimate, however, was conditional on a strong selection rule.
ChaosNLI re-annotated only items whose majority label carried exactly
three of the five original SNLI/MNLI annotator labels. Whether the monotonicity
boundary exists in the unselected populations those items came from was
never measured. This paper measures it, under a plan frozen before
analysis. We apply
the identical tagger, whose rules were frozen before analysis, to the SNLI
and MNLI development sets and test the planned prediction that
non-upward items sit lower on a four-level agreement ordinal built
from the original five labels.

The planned prediction is rejected. The sign reverses in all seven
contrasts, and every effect is small: the largest $\delta$ is $+0.059$,
far below our planned smallest effect size of interest (SESOI) of
0.10, and no confidence interval reaches it. The contribution of
this paper is the measured selection dependence itself. A boundary that
is present inside a disagreement-selected resource does not persist in
the unselected population.

\section{Related Work}
\label{sec:related}

\paragraph{NLI disagreement resources.} SNLI and MNLI development items
carry five labels each, the original writer's label plus four validation
labels \citep{bowman-etal-2015-large,williams-etal-2018-broad}; ChaosNLI
re-annotates a selected subset with 100 labels each
\citep[Section 3.1]{nie-etal-2020-learn}. The perspectivist literature
argues for modeling the resulting label distributions directly
\citep[Section 3]{plank-2022-problem}; see also
\citet{pavlick-kwiatkowski-2019-inherent}.

\paragraph{Monotonicity and NLI.} Monotonicity reasoning is a classic
locus of difficulty in NLI. \citet[Section 1.4.1]{maccartney-2009-natural}
notes that downward-entailing behavior is not confined to obvious
negation: ``constructions which are not upward monotone are surprisingly
widespread'', spanning quantifiers, conditional antecedents,
superlatives, and open-class triggers across several parts of speech.
The MED dataset targets
monotonicity reasoning directly \citep{yanaka-etal-2019-neural}; we use
it to anchor tagger error rates.

\paragraph{Operator semantics is itself contested.} For several operators
the monotonicity classification is a live semantic debate rather than a
lookup: \emph{only} has been analyzed as Strawson downward-entailing
rather than simply downward-entailing \citep{vonfintel-1999-npi}, and
\emph{many} has received both cardinal and proportional readings
\citep[Section 2.1.2]{rett-2018-semantics}; for that ambiguity, Rett
cites \citet{partee-1989-many} among others. This is relevant to the audit in
Section~\ref{sec:validity}: part of the tagger-human disagreement we
observe is irreducible for exactly this reason.

\section{Method}
\label{sec:method}

\paragraph{Data.} SNLI dev (9,986 items after excluding 14 items with
four validation labels) and MNLI matched dev (10,000 items). MNLI
mismatched dev (9,946 items after excluding 54 rows from 27 duplicated pair
IDs) is analyzed as a secondary corpus. The plan did not anticipate the
duplicate exclusion, and Appendix~\ref{app:concordance} records it as a
deviation. All corpora carry five labels per item. MED (5,382 premise-hypothesis pairs) is used only
for tagger validation, and the ChaosNLI overlap (1,514 SNLI rows, of which
1,507 remain after the same four-label exclusion, and 1,599 MNLI matched
items) only for the bridge analysis in Section~\ref{sec:results}.

\paragraph{Predictor.} A rule-based monotonicity operator tagger (v0.3,
frozen before this study under the plan; no rule changes) labels
each hypothesis upward, downward, non-monotone, or mixed by a surface
trigger scan. The analysis contrast is binary: purely upward versus
non-upward.

\paragraph{Outcome.} A four-level agreement ordinal from the five
validation labels: no majority $<$ 3/5 $<$ 4/5 $<$ 5/5. The primary
analysis includes the no-majority level; excluding it is sensitivity
analysis A.

\paragraph{Statistics.} Tie-corrected Cliff's $\delta$ with two-sided
Mann-Whitney U tests, percentile bootstrap CIs (10,000 resamples, seed
20260717), Holm correction over the two confirmatory contrasts (SNLI
primary, MNLI matched primary), and a planned SESOI of
$|\delta| = 0.10$. The planned prediction was $\delta < 0$ (non-upward
items lower in agreement). The plan was written into our working
repository on 2026-07-17 as an entry in its version-controlled research
log, before the confirmatory analysis was run. Its timestamp is a
version-control record under our own
control rather than a third-party registry entry, so we report the date
as a statement about our procedure and not as independently attested
evidence; \texttt{AUDIT.md} in the reproduction repository sets out the
plan's timeline in those terms, and the full
plan-versus-result concordance is in
Appendix~\ref{app:concordance}.

\paragraph{Bridge to the earlier estimate.} The originally planned
correlation of the ordinal outcome with ChaosNLI entropy inside the
overlap is undefined by arithmetic: ChaosNLI selected only items with a 3/5 modal
label, so the agreement ordinal is constant (variance zero) on the entire
overlap. We state this openly and replace the bridge with level-wise
descriptive statistics (Section~\ref{sec:results}). Scale caveat: the
earlier $\delta$ was computed on a 100-label entropy outcome and is not
numerically commensurable with $\delta$ values on the four-level ordinal;
we compare sign and magnitude class only.

\section{Results}
\label{sec:results}

\begin{table*}[t]
\centering
\small
\setlength{\tabcolsep}{4pt}
\begin{tabular}{llccll}
\toprule
Contrast & Status & $n$ non-up / up & $\delta$ & 95\% CI & $p$ \\
\midrule
SNLI primary & confirmatory & 551 / 9,435 & $+0.032$ & $[-0.011, +0.075]$ & 0.162 (Holm) \\
MNLI matched primary & confirmatory & 3,329 / 6,671 & $+0.045$ & $[+0.024, +0.066]$ & $7.1 \times 10^{-5}$ (Holm) \\
MNLI mismatched & secondary & 3,402 / 6,544 & $+0.059$ & $[+0.038, +0.079]$ & $4.1 \times 10^{-8}$ \\
SNLI, no-majority dropped & sensitivity A & 544 / 9,287 & $+0.030$ & $[-0.014, +0.072]$ & 0.193 \\
Matched, no-majority dropped & sensitivity A & 3,273 / 6,542 & $+0.044$ & $[+0.023, +0.065]$ & $5.3 \times 10^{-5}$ \\
Mismatched, no-majority dropped & sensitivity A & 3,351 / 6,427 & $+0.058$ & $[+0.036, +0.078]$ & $6.9 \times 10^{-8}$ \\
SNLI + matched merged & sensitivity B & 3,880 / 16,106 & $+0.045$ & $[+0.027, +0.063]$ & $1.0 \times 10^{-6}$ \\
\bottomrule
\end{tabular}
\caption{All contrasts. Planned prediction: $\delta < 0$. SESOI:
$|\delta| = 0.10$. No CI reaches the SESOI boundary. The earlier interval
$[-0.332, -0.235]$ sits on a non-commensurable outcome scale, so
comparison with it is one of sign and magnitude class only.}
\label{tab:main}
\end{table*}

\paragraph{The planned prediction is rejected.} Table~\ref{tab:main}
reports all seven contrasts. Every $\delta$ is positive: non-upward items
sit very slightly higher, not lower, on the agreement ordinal. The SNLI
primary contrast is not significant ($\delta = +0.032$, 95\% CI
$[-0.011, +0.075]$, Holm $p = 0.162$). The MNLI matched primary contrast
is significant but with the sign opposite to the plan
($\delta = +0.045$, 95\% CI $[+0.024, +0.066]$, Holm
$p = 7.1 \times 10^{-5}$). The secondary mismatched corpus behaves like
matched ($\delta = +0.059$). Dropping no-majority items (sensitivity A)
and merging SNLI with matched (sensitivity B) change nothing material.

\begin{figure}[t]
\centering
\includegraphics[width=\columnwidth]{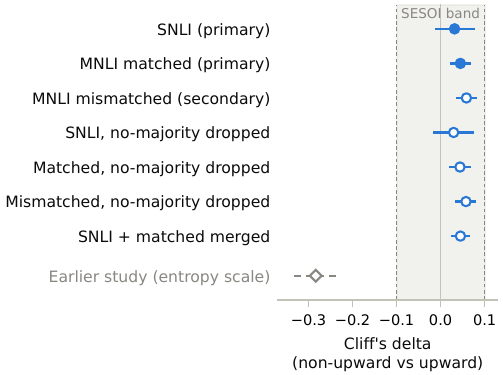}
\caption{The seven contrasts of Table~\ref{tab:main}. Points are
tie-corrected Cliff's $\delta$ with percentile bootstrap 95\% confidence
intervals; the shaded band marks the planned SESOI at
$|\delta| = 0.10$. The gray reference row is labeled \emph{Earlier
study (entropy scale)} and reports that study's estimate. It is
computed on a 100-label
entropy outcome that is not numerically commensurable with the
four-level ordinal used here and is shown for sign and magnitude
comparison only. Generated by
\texttt{analysis/fig\_forest.py}.}
\label{fig:forest}
\end{figure}

\paragraph{Effects are below the SESOI everywhere.} All
$|\delta| < 0.10$, and no bootstrap CI touches the boundary
(Figure~\ref{fig:forest}). Item-level
discriminability is essentially absent: ordered logistic pseudo-$R^2$ is
0.000093 (SNLI) and 0.00084 (matched), and the AUC for separating
high-agreement (4/5, 5/5) from low-agreement items is at chance in both
corpora (0.492 SNLI, 0.484 matched).

\paragraph{Outcome resolution does not explain the reversal.} The earlier
study read disagreement off 100 labels per item, while the ordinal used
here has five levels, so one might ask whether the boundary failed to
replicate only because the outcome lost resolution. Coarsening an outcome
generally attenuates an association toward zero rather than reversing its
sign, and
every $\delta$ we observe is positive, so lost resolution cannot
manufacture the direction we report. What resolution does bound is how
precisely the two studies can be compared, which is why we restrict that
comparison to sign and magnitude class. Subsampling experiments that vary
annotation count from one to one hundred labels per item on ChaosNLI make
that dependence explicit \citep{kadasi-singh-2023-unveiling}.

\paragraph{Complexity confounders do not explain the reversal.} As an
auxiliary, non-confirmatory defense we fit proportional-odds models of
the agreement ordinal on the binary monotonicity predictor, without (M0)
and with (M1) length, parse depth, and genre covariates. In matched and
mismatched the monotonicity coefficient keeps its sign and stays away
from zero after adjustment (matched M1: $-0.202$,
$p = 2.4 \times 10^{-6}$; mismatched M1: $-0.248$,
$p = 7.3 \times 10^{-9}$; negative coefficients here correspond to the
positive $\delta$ values above, because upward items sit lower in
agreement). In SNLI the proportional-odds assumption fails a Brant test
\citep{brant-1990-assessing}, and the
multinomial substitute shows no significant category-wise monotonicity
effect, which is consistent with the non-significant SNLI $\delta$.

\begin{figure}[t]
\centering
\includegraphics[width=\columnwidth]{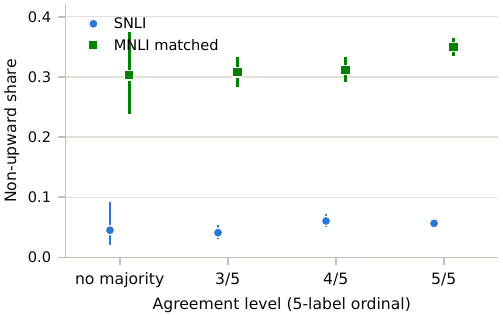}
\caption{Share of non-upward hypotheses at each agreement level, with
Wilson 95\% confidence intervals, for SNLI dev and MNLI matched dev. The
share does not decrease as agreement rises. Generated by
\texttt{analysis/fig\_bridge\_shares.py}.}
\label{fig:bridge}
\end{figure}

\paragraph{Level-wise structure runs against the planned direction.}
The share of non-upward hypotheses does not decrease with agreement
level. In matched it is highest among unanimous items: 0.303 (no
majority), 0.308 (3/5), 0.312 (4/5), 0.350 (5/5). SNLI shows a flat low
profile (0.045, 0.041, 0.060, 0.057). Figure~\ref{fig:bridge} plots both
corpora; Appendix~\ref{app:bridge} gives the arithmetic.

\section{Validity of the Measurement}
\label{sec:validity}

\paragraph{Tagger misclassification (Tier 2).} We anchored flip rates in
MED measurements of the tagger (symmetric error $542/5{,}382 = 0.101$;
upward to non-upward $229/1{,}818 = 0.126$; non-upward to upward
$313/3{,}564 = 0.088$; denominators recomputed from the released MED
file, which differs by two items from the class counts reported in
\citealp{yanaka-etal-2019-neural}; MED domain, not a dev-domain error
claim) and
simulated label flips over a 36-cell grid (three corpora, three
scenarios, anchored plus fixed rates, 1,000 replications each).
Misclassification shrinks $\delta$ toward zero rather than manufacturing
it: at the MED-anchored symmetric rate the matched $\delta$ averages
0.0345 with significance retained in 94.9\% of replications, and
mismatched retains it in every replication.
Retention drops only under the extreme fixed rates, most steeply for
matched at symmetric 0.20 (0.594), where mismatched still holds at
0.933. Across all 36
cells the 97.5th percentile of simulated $\delta$ never exceeds 0.064, so
the below-SESOI conclusion never flips (Appendix~\ref{app:tier2}).

\paragraph{Manual audit (Tier 3).} We hand-tagged a fresh blind sample of
200 hypotheses (50 each from SNLI dev, MNLI matched dev, and the two
ChaosNLI overlap strata) against a written codebook that fixes the
judgment target: does the hypothesis contain a monotonicity trigger,
judged by semantic definition rather than a word list. Four-class
agreement with the tagger was 0.875 (Wilson 95\% CI $[0.822, 0.914]$),
binary agreement 0.905, Cohen's $\kappa$ 0.607 (four-class; both
marginals are heavily skewed toward upward, so $\kappa$ understates
agreement and is read alongside the raw rates). An earlier round without
a codebook produced agreement of 0.135. The gap is an
instrument-specification result rather than an error: without a fixed
judgment target, the human rater measured a different construct (premise-relative
entailment direction rather than trigger presence). We report both
rounds.

\paragraph{Correlated misclassification.} The flips simulated in Tier 2
are drawn independently of the agreement outcome, so that tier bounds
random misclassification only. Error correlated with the outcome can move
$\delta$ in a direction instead of toward zero. We estimated that
component on the 100 Tier 3 items that come from the two confirmatory dev
cells, recomputing $\delta$ on those same items once under the manual
labels and once under the tagger labels. The manual estimate sits 0.026
below the tagger estimate (95\% CI $[-0.288, +0.230]$; 20,000 item-level
resamples, seed 20260726), so the tagger displaces $\delta$ toward the
positive sign we observe, which is the direction a correlated error would
produce. The interval covers zero and is an order of magnitude wider than
the displacement, so 100 items settle neither the size nor the sign, and
we give the point estimate only for scale. A displacement of that scale
leaves every confirmatory contrast positive and below SESOI, so we report
it as a bound on the error model rather than as a correction.
\citet{choi-2026-hlv} records the complementary coverage gap from the
other side, reporting that the same tagger does not process conditional
and comparative environments, which occur in 7.4\% of that study's items,
and leaving the direction of full coverage unresolved.

\paragraph{Three layers of residual disagreement.} We read the 25
disagreements by hand (Appendix~\ref{app:tier3}) and assigned each to
an error layer. This layer
coding is a qualitative judgment, not a script output; the counts below
are approximate and are the one set of numbers in this paper that no
script regenerates. They decompose into (a) about 10 tagger lexicon
coverage gaps, (b) about 5
polysemy and construction failures, and (c) about 10 items where the
operator's monotonicity classification is itself semantically contested,
including \emph{only} (rated downward by the human 4 times and left
upward 3 times within this sample; \citealp{vonfintel-1999-npi}) and
\emph{many} \citep{rett-2018-semantics}. Layers (a) and (b) are
engineering-reducible; we judge layer (c) not to be, and it is itself
direct evidence for treating disagreement as a measurement target. We
deliberately did not patch the tagger in response to this audit, because
reacting to the audit sample would circularize the validation.

\section{Discussion}
\label{sec:discussion}

The leading interpretation is that the earlier boundary is a structure
conditional on low-agreement selection. Inside a resource restricted to
3/5-split items, non-upward operators separated items by residual
disagreement; in the unselected population, they do not. The residual
associations all carry the opposite sign, but none of them reaches the
SESOI, so we do not read them as a reversal. The immediate implication is
that HLV structure claims estimated on selected re-annotation resources
should carry their selection conditional explicitly; the ChaosNLI
selection rule is strong enough to zero out the variance of the four-level
agreement ordinal on the overlap, which is why the natural bridge
analysis is undefined by arithmetic.

A second observation concerns where disagreement reappeared in this
study. About 10 of the 25 tagger-human disagreements are items whose
operator classification is contested in the semantics literature, not
items on which the tagger is simply wrong. Within this sample
\emph{only} was rated downward 4 times and left upward 3 times, and that
split tracks a question the literature has not settled
\citep{vonfintel-1999-npi}; \emph{many} carries a comparable open status
\citep{rett-2018-semantics}. The phenomenon the study takes as its object
therefore recurs inside the instrument built to measure it. We read this
as a property of the construct rather than as a defect of this particular
tagger, since any instrument that sorts items into monotonicity classes
has to resolve the same contested cases, and lexicon engineering does not
reach them. Measurement frames that treat annotator disagreement as noise
to be eliminated would misread this layer specifically, because here the
disagreement sits in the analytic vocabulary itself.

The plan written before analysis, the SESOI, and the freeze discipline
are what give the negative result its evidential value; we report it as
the plan specified.

\section{Reproducibility}
\label{sec:repro}

Every quantity this paper reports is printed in the paper itself,
including the full appendix tables, so no claim requires external
material. The reproduction repository (\repourl) supplies the code that
regenerates those quantities. File paths named here are relative to that
repository's root.

All reported numbers regenerate from the analysis scripts, except the
manual error layer counts in Section~\ref{sec:validity}, which come from
manual inspection of the items in Table~\ref{tab:d2}. One further
quantity sits outside the planned pipeline. The correlated
misclassification estimate in Section~\ref{sec:validity} was computed
after the planned analyses closed, and it regenerates from
\texttt{analysis/exploratory/\allowbreak directional\_bias\_probe.py}, which is kept in a
separate directory precisely so that nothing in the plan can depend on it. Every run that involves randomness
fixes a seed and records it in that run's output file. The repository
README maps each table, figure, and statistic to the step that produces
it, and gives the run order.

\texttt{AUDIT.md} is the audit trail. It records the plan's
timeline, what had already been observed at each point where the plan
was extended, every deviation from the plan, and every negative result. The
repository is a curated snapshot, not a full development history, and
\texttt{AUDIT.md} states what that costs.

\section*{Limitations}

Manual tagging was performed by a single author, blind to tagger output
but without an independent second rater; the reported $\kappa$ is
tagger-versus-human, not inter-human. The tagger is a surface trigger
scanner with measured coverage gaps (layers (a) and (b) of
Section~\ref{sec:validity}). MED anchor rates are measured on MED
sentences, not on the SNLI/MNLI dev domain. The four-level agreement
ordinal is coarse, and all data are English. The earlier study and this
replication use non-commensurable outcome scales, so the replication
verdict rests on sign and magnitude class, not on interval overlap.


\bibliography{refs}

\appendix

\section{Plan-versus-Result Concordance}
\label{app:concordance}

The plan is section 1 of the document written into the working
repository and frozen on 2026-07-17; the item numbers below refer to it,
and the Plan column reproduces what each item specified, so the
concordance can be read without consulting that document. Item 1.3 of
the plan covers both the predictor freeze and the confirmatory
hypothesis, so it occupies two rows. Status
values: ``as
planned'' (followed without change), ``prediction rejected''
(procedure followed, planned direction failed), ``documented
deviation'' (change made during the study, recorded in the analysis log
before or at the point of change). Tables~\ref{tab:concordance-a}
and~\ref{tab:concordance-b} give the full concordance.

\begin{table*}[t]
\centering
\footnotesize
\begin{tabular}{p{0.18\textwidth}p{0.33\textwidth}p{0.41\textwidth}}
\toprule
Planned item & Plan & Status \\
\midrule
Outcome variable (1.1) & Four-level agreement ordinal, no-majority as
lowest level, primary; three-level exclusion as sensitivity & as
planned \\
\addlinespace
Sample and exclusions (1.2) & SNLI dev minus 14 four-label items; MNLI
matched in full; no other exclusions & as planned for confirmatory
corpora; documented deviation for the secondary corpus (54 rows from 27
duplicated pair IDs in mismatched, excluded with logged rationale) \\
\addlinespace
Predictor freeze (1.3) & Tagger v0.3 reused with no rule changes & as
planned; the Tier 3 audit did not trigger modification, by explicit
decision \\
\addlinespace
Confirmatory hypothesis (1.3) & Non-upward group sits lower on the
agreement ordinal & prediction rejected; all seven contrasts positive
(Table~\ref{tab:main}) \\
\addlinespace
Statistics (1.4) & Tie-corrected Cliff's $\delta$, two-sided
Mann-Whitney U, percentile bootstrap 10,000 draws, seed 20260717, Holm
over the two confirmatory contrasts & as planned \\
\addlinespace
Confound defense (1.4) & Proportional-odds model; multinomial substitute
on assumption violation, violation reported & as planned; the
violation occurred in SNLI and is reported in
Section~\ref{sec:results} \\
\bottomrule
\end{tabular}
\caption{Plan-versus-result concordance, planned items 1.1 to
1.4.}
\label{tab:concordance-a}
\end{table*}

\begin{table*}[t]
\centering
\footnotesize
\begin{tabular}{p{0.18\textwidth}p{0.33\textwidth}p{0.41\textwidth}}
\toprule
Planned item & Plan & Status \\
\midrule
Power and SESOI (1.5) & SESOI at absolute $\delta$ 0.10; report CI
position against the boundary & as planned; every estimate is below
the SESOI and no CI touches the boundary \\
\addlinespace
Secondary and sensitivity analyses (1.6) & Mismatched pipeline,
no-majority exclusion, merged corpus, outside the confirmatory family &
as planned \\
\addlinespace
Item-level ceiling (1.7) & Ordinal pseudo-$R^2$ and high-versus-low AUC
with a noncommensurability caution & as planned
(Section~\ref{sec:results}) \\
\addlinespace
ChaosNLI bridge (1.8) & Three-component redefinition after the original
correlation design was found undefined & as planned
(Section~\ref{sec:results} and Appendix~\ref{app:bridge}) \\
\addlinespace
Tier 2 simulation (1.9) & Flip grid over fixed rates plus MED-anchored
rates, 1,000 replications, seed 20260717 & as planned
(Appendix~\ref{app:tier2}) \\
\addlinespace
Tier 3 manual audit (1.10) & 200-item blind manual tagging, four cells of
50, single-author limitation stated & documented deviation: the first
round lacked a written codebook and measured a different construct
(agreement 0.135); a codebook was then written and a fresh 200-item
sample re-tagged, with the original round excluded from the headline
metric and both rounds reported (Section~\ref{sec:validity}) \\
\bottomrule
\end{tabular}
\caption{Plan-versus-result concordance, planned items 1.5 to
1.10.}
\label{tab:concordance-b}
\end{table*}

\section{Bridge Details and Overlap Arithmetic}
\label{app:bridge}

\input{appendix_tables_b}

The original bridge design (correlating five-label agreement with
ChaosNLI 100-label entropy on the overlap) is undefined by arithmetic:
ChaosNLI selected items whose majority label carries exactly three votes,
so the agreement ordinal has zero variance on the overlap. The overlap
arithmetic confirms this exactly. All 1,599 ChaosNLI-M rows match MNLI
matched dev items at level 3/5, and the 1,514 ChaosNLI-S rows decompose
into 1,507 five-label items at 3/5 plus 7 four-label items at 3/4; the
latter fall under the planned four-label exclusion, leaving 1,507
SNLI bridge items (Table~\ref{tab:b1}).

The planned replacement has three components. Component 1 reports the
spread of ChaosNLI 100-label entropy inside the constant-ordinal overlap,
measuring the coarseness of the ordinal (Table~\ref{tab:b2}): the SNLI overlap has
mean entropy 0.800 (sd 0.332) and the matched overlap 1.072 (sd 0.267),
so substantial graded variation survives within a single ordinal level.
Component 2 reports level-wise non-upward shares with Wilson CIs
(Table~\ref{tab:b3}, plotted as Figure~\ref{fig:bridge}). Component 3 is
this transparency statement itself.

\section{Full Tier 2 Grid}
\label{app:tier2}

\input{appendix_tables_c}

Table~\ref{tab:c} lists all 36 cells: three corpora,
three flip scenarios (symmetric, upward-to-non-upward only, reverse), and
four rates per scenario arm (fixed 0.05, 0.10, 0.20 plus the MED-anchored
rate for that scenario), with 1,000 replications per cell at seed
20260717. Reported per cell: mean simulated $\delta$, the 2.5th and
97.5th percentiles, and significance retention. The maximum 97.5th
percentile across all cells is 0.064, which is the basis for the claim in
Section~\ref{sec:validity} that the below-SESOI conclusion does not flip
under simulated misclassification.

\section{Tier 3 Codebook Summary and Disagreements}
\label{app:tier3}

\input{appendix_tables_d}

The codebook (v1, 2026-07-19) fixes the judgment target: whether the hypothesis sentence contains a
monotonicity trigger, judged by semantic definition (does the expression
license or block upward or downward substitution in its scope) rather
than by membership in a word list. It defines the four-class label set
(upward, downward, non-monotone, mixed), an existence-based decision
rule, boundary-case guidance, and a circularity note: the codebook was
written from semantic definitions, not from the tagger's rule inventory,
so agreement with the tagger is not built in.

Table~\ref{tab:d1} reports the full agreement metrics for the codebook
round, including per-type disagreement counts. Table~\ref{tab:d2} lists
all 25 disagreement items verbatim (cell, manual label, tagger label,
detected triggers, hypothesis text), grouped by disagreement type. These
25 items are the basis for the three-layer decomposition in
Section~\ref{sec:validity}.

\end{document}

%% file: appendix_tables_b.tex
%

\begin{table}[htbp]
\centering
\footnotesize
\begin{tabular}[t]{lr}
\toprule
Metric & Value \\
\midrule
\multicolumn{2}{l}{\itshape SNLI} \\
chaos rows & 1514 \\
dev rows & 10000 \\
matched total & 1514 \\
matched 5-label & 1507 \\
matched 4-label & 7 \\
unmatched & 0 \\
agreement key 3/5 & 1507 \\
agreement key 3/4 & 7 \\
\midrule
\multicolumn{2}{l}{\itshape MNLI matched} \\
chaos rows & 1599 \\
dev rows & 10000 \\
matched total & 1599 \\
matched 5-label & 1599 \\
matched 4-label & 0 \\
unmatched & 0 \\
agreement key 3/5 & 1599 \\
\bottomrule
\end{tabular}
\caption{ChaosNLI overlap arithmetic. Every matched item falls at a single agreement level, which is why the planned bridge correlation is undefined: the agreement ordinal has zero variance on the overlap.}
\label{tab:b1}
\end{table}

\begin{table*}[tp]
\centering
\footnotesize
\begin{tabular}[t]{lrrrrrrrr}
\toprule
Corpus & $n$ & Mean & SD & Median & Q1 & Q3 & Min & Max \\
\midrule
SNLI & 1507 & 0.800 & 0.332 & 0.843 & 0.579 & 1.005 & 0.000 & 1.583 \\
MNLI matched & 1599 & 1.072 & 0.267 & 1.089 & 0.925 & 1.252 & 0.081 & 1.584 \\
\bottomrule
\end{tabular}
\caption{ChaosNLI 100-label entropy inside the constant-ordinal overlap. Substantial graded variation survives within a single agreement level, which is the roughness the ordinal cannot express.}
\label{tab:b2}
\end{table*}

\begin{table}[htbp]
\centering
\footnotesize
\begin{tabular}[t]{lrrrr}
\toprule
Level & $n$ & Share & CI lo & CI hi \\
\midrule
\multicolumn{5}{l}{\itshape SNLI} \\
nomaj/5 & 155 & 0.045 & 0.022 & 0.090 \\
3/5 & 1507 & 0.041 & 0.032 & 0.052 \\
4/5 & 2845 & 0.060 & 0.052 & 0.070 \\
5/5 & 5479 & 0.057 & 0.051 & 0.063 \\
\midrule
\multicolumn{5}{l}{\itshape MNLI matched} \\
nomaj/5 & 185 & 0.303 & 0.241 & 0.372 \\
3/5 & 1599 & 0.308 & 0.286 & 0.331 \\
4/5 & 2457 & 0.312 & 0.294 & 0.330 \\
5/5 & 5759 & 0.350 & 0.338 & 0.362 \\
\bottomrule
\end{tabular}
\caption{Share of non-upward hypotheses at each agreement level, with Wilson 95\% confidence intervals. The share does not decrease as agreement rises. Plotted as Figure~\ref{fig:bridge}.}
\label{tab:b3}
\end{table}

%% file: appendix_tables_c.tex
%

\begin{table*}[tp]
\centering
\scriptsize
\begin{tabular}[t]{llrrrrrr}
\toprule
Scenario & Rate & $n$ valid & $n$ degen. & Mean $\delta$ & 2.5th pct & 97.5th pct & Sig. ret. \\
\midrule
\multicolumn{8}{l}{\itshape MNLI matched} \\
directional\_non\_to\_up & 0.0500 & 1000 & 0 & 0.0439 & 0.0381 & 0.0492 & 1.000 \\
directional\_non\_to\_up & 0.0878 (anchor) & 1000 & 0 & 0.0430 & 0.0354 & 0.0510 & 1.000 \\
directional\_non\_to\_up & 0.1000 & 1000 & 0 & 0.0430 & 0.0352 & 0.0514 & 1.000 \\
directional\_non\_to\_up & 0.2000 & 1000 & 0 & 0.0409 & 0.0287 & 0.0529 & 0.990 \\
directional\_up\_to\_non & 0.0500 & 1000 & 0 & 0.0407 & 0.0334 & 0.0483 & 1.000 \\
directional\_up\_to\_non & 0.1000 & 1000 & 0 & 0.0373 & 0.0271 & 0.0471 & 0.995 \\
directional\_up\_to\_non & 0.1260 (anchor) & 1000 & 0 & 0.0362 & 0.0258 & 0.0463 & 0.992 \\
directional\_up\_to\_non & 0.2000 & 1000 & 0 & 0.0320 & 0.0188 & 0.0454 & 0.909 \\
symmetric & 0.0500 & 1000 & 0 & 0.0396 & 0.0306 & 0.0501 & 1.000 \\
symmetric & 0.1000 & 1000 & 0 & 0.0345 & 0.0218 & 0.0470 & 0.946 \\
symmetric & 0.1007 (anchor) & 1000 & 0 & 0.0345 & 0.0220 & 0.0465 & 0.949 \\
symmetric & 0.2000 & 1000 & 0 & 0.0249 & 0.0074 & 0.0398 & 0.594 \\
\midrule
\multicolumn{8}{l}{\itshape MNLI mismatched} \\
directional\_non\_to\_up & 0.0500 & 1000 & 0 & 0.0573 & 0.0514 & 0.0631 & 1.000 \\
directional\_non\_to\_up & 0.0878 (anchor) & 1000 & 0 & 0.0562 & 0.0486 & 0.0635 & 1.000 \\
directional\_non\_to\_up & 0.1000 & 1000 & 0 & 0.0557 & 0.0477 & 0.0633 & 1.000 \\
directional\_non\_to\_up & 0.2000 & 1000 & 0 & 0.0531 & 0.0413 & 0.0639 & 1.000 \\
directional\_up\_to\_non & 0.0500 & 1000 & 0 & 0.0537 & 0.0459 & 0.0607 & 1.000 \\
directional\_up\_to\_non & 0.1000 & 1000 & 0 & 0.0490 & 0.0391 & 0.0588 & 1.000 \\
directional\_up\_to\_non & 0.1260 (anchor) & 1000 & 0 & 0.0473 & 0.0355 & 0.0581 & 1.000 \\
directional\_up\_to\_non & 0.2000 & 1000 & 0 & 0.0423 & 0.0284 & 0.0550 & 0.998 \\
symmetric & 0.0500 & 1000 & 0 & 0.0518 & 0.0421 & 0.0607 & 1.000 \\
symmetric & 0.1000 & 1000 & 0 & 0.0454 & 0.0320 & 0.0581 & 1.000 \\
symmetric & 0.1007 (anchor) & 1000 & 0 & 0.0450 & 0.0322 & 0.0577 & 1.000 \\
symmetric & 0.2000 & 1000 & 0 & 0.0330 & 0.0164 & 0.0499 & 0.933 \\
\midrule
\multicolumn{8}{l}{\itshape SNLI} \\
directional\_non\_to\_up & 0.0500 & 1000 & 0 & 0.0317 & 0.0216 & 0.0413 & 0.000 \\
directional\_non\_to\_up & 0.0878 (anchor) & 1000 & 0 & 0.0317 & 0.0190 & 0.0454 & 0.001 \\
directional\_non\_to\_up & 0.1000 & 1000 & 0 & 0.0319 & 0.0166 & 0.0465 & 0.000 \\
directional\_non\_to\_up & 0.2000 & 1000 & 0 & 0.0317 & 0.0091 & 0.0547 & 0.016 \\
directional\_up\_to\_non & 0.0500 & 1000 & 0 & 0.0172 & -0.0065 & 0.0413 & 0.038 \\
directional\_up\_to\_non & 0.1000 & 1000 & 0 & 0.0120 & -0.0108 & 0.0342 & 0.037 \\
directional\_up\_to\_non & 0.1260 (anchor) & 1000 & 0 & 0.0103 & -0.0120 & 0.0340 & 0.044 \\
directional\_up\_to\_non & 0.2000 & 1000 & 0 & 0.0070 & -0.0135 & 0.0270 & 0.025 \\
symmetric & 0.0500 & 1000 & 0 & 0.0161 & -0.0080 & 0.0398 & 0.030 \\
symmetric & 0.1000 & 1000 & 0 & 0.0108 & -0.0131 & 0.0353 & 0.041 \\
symmetric & 0.1007 (anchor) & 1000 & 0 & 0.0108 & -0.0147 & 0.0365 & 0.045 \\
symmetric & 0.2000 & 1000 & 0 & 0.0050 & -0.0181 & 0.0270 & 0.026 \\
\bottomrule
\end{tabular}
\caption{Tier 2 misclassification grid, all 36 cells. Seed 20260717, 1000 replications per cell. Rates marked (anchor) are the MED-anchored rate for that scenario. The maximum 97.5th percentile across all cells is 0.064, below the planned SESOI of $|\delta| = 0.10$.}
\label{tab:c}
\end{table*}

%% file: appendix_tables_d.tex
%

\begin{table*}[tp]
\centering
\footnotesize
\begin{tabular}[t]{lr}
\toprule
Metric & Value \\
\midrule
n\_items & 200 \\
four-class agreement rate & 0.875 \\
four-class agreement n\_agree & 175 \\
four-class agreement n\_total & 200 \\
four-class agreement wilson ci lo & 0.822 \\
four-class agreement wilson ci hi & 0.914 \\
binary agreement rate & 0.905 \\
binary agreement n\_agree & 181 \\
binary agreement n\_total & 200 \\
binary agreement wilson ci lo & 0.856 \\
binary agreement wilson ci hi & 0.938 \\
four-class kappa & 0.607 \\
binary kappa & 0.688 \\
\bottomrule
\end{tabular}
\hfill
\begin{tabular}[t]{lr}
\toprule
Metric & Value \\
\midrule
disagreement type downward -> upward & 7 \\
disagreement type upward -> non\_monotone & 6 \\
disagreement type downward -> non\_monotone & 3 \\
disagreement type upward -> downward & 3 \\
disagreement type mixed -> downward & 2 \\
disagreement type mixed -> upward & 2 \\
disagreement type downward -> mixed & 1 \\
disagreement type non\_monotone -> upward & 1 \\
disagreement type mixed -> non\_monotone & 0 \\
disagreement type non\_monotone -> downward & 0 \\
disagreement type non\_monotone -> mixed & 0 \\
disagreement type upward -> mixed & 0 \\
\bottomrule
\end{tabular}
\caption{Tier 3 agreement between the manual codebook labels and the tagger, with per-type disagreement counts. Directions read manual to tagger. Agreement is tagger versus human, not inter-human. The rows run down the left panel and continue in the right.}
\label{tab:d1}
\end{table*}

\begin{table*}[tp]
\centering
\scriptsize
\setlength{\tabcolsep}{4pt}
\begin{tabular}[t]{@{}rlllp{0.20\textwidth}p{0.38\textwidth}@{}}
\toprule
Item & Cell & Manual & Tagger & Triggers & Hypothesis \\
\midrule
\multicolumn{6}{l}{\itshape downward -> upward (7 items)} \\
7 & chaosnli\_s & downward & upward & (none detected) & 50 people walked to the wrong subway hall. \\
34 & chaosnli\_m & downward & upward & (none detected) & It is better to plant when it is colder. \\
38 & mnli\_matched & downward & upward & (none detected) & It's impossible to have a plate hand-painted to your own design in Hong Kong. \\
47 & mnli\_matched & downward & upward & (none detected) & European members of NATO might consider the US's efforts to be less credible. \\
143 & chaosnli\_m & downward & upward & (none detected) & Was it Jane Eyre or not? \\
146 & chaosnli\_m & downward & upward & (none detected) & They would get upset whenever anyone would speak to them. \\
173 & mnli\_matched & downward & upward & (none detected) & We went to the office to see if there was anything we could rent. \\
\midrule
\multicolumn{6}{l}{\itshape upward -> non\_monotone (6 items)} \\
21 & mnli\_matched & upward & non\_monotone & only (non\_monotone) & Zelon is the only student-run chapter of the American Civil Liberties Union in New York state. \\
58 & chaosnli\_m & upward & non\_monotone & only (non\_monotone) & Lister and Simpson were the only ones to use carbolic acid and chloroform for this purpose. \\
121 & mnli\_matched & upward & non\_monotone & last (non\_monotone) & Your mistress wrote letters last night. \\
128 & chaosnli\_m & upward & non\_monotone & many (non\_monotone) & Our higher-end stores have been suffering due to the recession, and many have shut down for lack of revenue. \\
184 & chaosnli\_s & upward & non\_monotone & only (non\_monotone) & There are only two people in the field. \\
188 & mnli\_matched & upward & non\_monotone & last (non\_monotone) & It was 37 degrees last night. \\
\midrule
\multicolumn{6}{l}{\itshape downward -> non\_monotone (3 items)} \\
24 & chaosnli\_m & downward & non\_monotone & only (non\_monotone) & There were only a few villas the whole way along, until we reached a small village that seemed to be the end. \\
95 & mnli\_matched & downward & non\_monotone & only (non\_monotone) & Price hikes are only possible if there is more money in circulation. \\
119 & mnli\_matched & downward & non\_monotone & smallest (non\_monotone), only (non\_monotone) & The library is the smallest estate in Jamaica, with only three books. \\
\midrule
\multicolumn{6}{l}{\itshape upward -> downward (3 items)} \\
3 & chaosnli\_s & upward & downward & each (downward) & Two dogs chase each other in the high grass. \\
66 & chaosnli\_m & upward & downward & n't (downward) & I wish you hadn't revealed your identity, that was a mistake. \\
181 & snli\_dev & upward & downward & little (downward) & A little girl is blowing the petals. \\
\midrule
\multicolumn{6}{l}{\itshape mixed -> downward (2 items)} \\
82 & mnli\_matched & mixed & downward & all (downward) & All of the homes in the hillside have been converted into art galleries and shops selling collectibles. \\
112 & mnli\_matched & mixed & downward & n't (downward) & I chose to become an actor, but I wasn't very good at it. \\
\midrule
\multicolumn{6}{l}{\itshape mixed -> upward (2 items)} \\
65 & mnli\_matched & mixed & upward & (none detected) & Everything can be found inside a shopping mall. \\
111 & chaosnli\_s & mixed & upward & (none detected) & The men are higher than the wall. \\
\midrule
\multicolumn{6}{l}{\itshape downward -> mixed (1 items)} \\
133 & chaosnli\_m & downward & mixed & only (non\_monotone), barely (downward) & The tunnel of Eupalinos is only one foot in diameter, barely large enough for a child to squeeze through. \\
\midrule
\multicolumn{6}{l}{\itshape non\_monotone -> upward (1 items)} \\
166 & chaosnli\_m & non\_monotone & upward & (none detected) & Although it was unnecessary, some of the equipment was adjacent. \\
\bottomrule
\end{tabular}
\caption{All 25 Tier 3 disagreement items, grouped by disagreement type. Group headings read manual label to tagger label. These items are the basis for the three-layer decomposition in Section~\ref{sec:validity}.}
\label{tab:d2}
\end{table*}